\documentclass[11pt,a4paper]{article}
\usepackage[hyperref]{acl2018}
\usepackage{times}
\usepackage{latexsym}
\usepackage{url}

\usepackage{comment}
\usepackage{array}
\usepackage{amsmath}
\usepackage[]{algorithm2e}
\usepackage{algpseudocode}
\usepackage{float}
\usepackage{graphicx}
\usepackage{color}
\definecolor{myblue}{RGB}{17, 118, 183}
\definecolor{myred}{RGB}{234, 38, 30}

\aclfinalcopy 
\def\aclpaperid{1612} 

\title{Learning to Write with Cooperative Discriminators}

\author{\quad \: Ari Holtzman$^\dagger$  \quad \quad \quad \: Jan Buys$^\dagger$ \quad \quad \quad Maxwell Forbes$^\dagger$ \\
\bf Antoine Bosselut$^\dagger$ \quad  \quad  David Golub$^\dagger$ \quad \quad \quad Yejin Choi$^\dagger$$^\ddagger$ \quad  \\
$^\dagger$Paul G. Allen School of Computer Science \& Engineering, University of Washington\\
$^\ddagger$Allen Institute for Artificial Intelligence\\
\texttt{\{ahai,jbuys,mbforbes,antoineb,golubd,yejin\}@cs.washington.edu}
}

\date{}

\begin{document}
\maketitle

\begin{abstract}
Despite their local fluency, long-form text generated from RNNs is often generic, repetitive, and even self-contradictory. 
We propose a unified learning framework that collectively addresses all the above issues by composing a committee of discriminators that can guide a base RNN generator towards more globally coherent generations.
More concretely, discriminators each specialize in a different principle of communication, such as Grice's maxims, and are collectively combined with the base RNN generator through a composite decoding objective.  
Human evaluation demonstrates that text generated by our model is preferred over that of baselines by a large margin, significantly enhancing the overall coherence, style, and information of the generations.
\end{abstract}


\section{Introduction}

\begin{figure}[t!]
\vspace*{-2mm}
\includegraphics[width=\columnwidth]{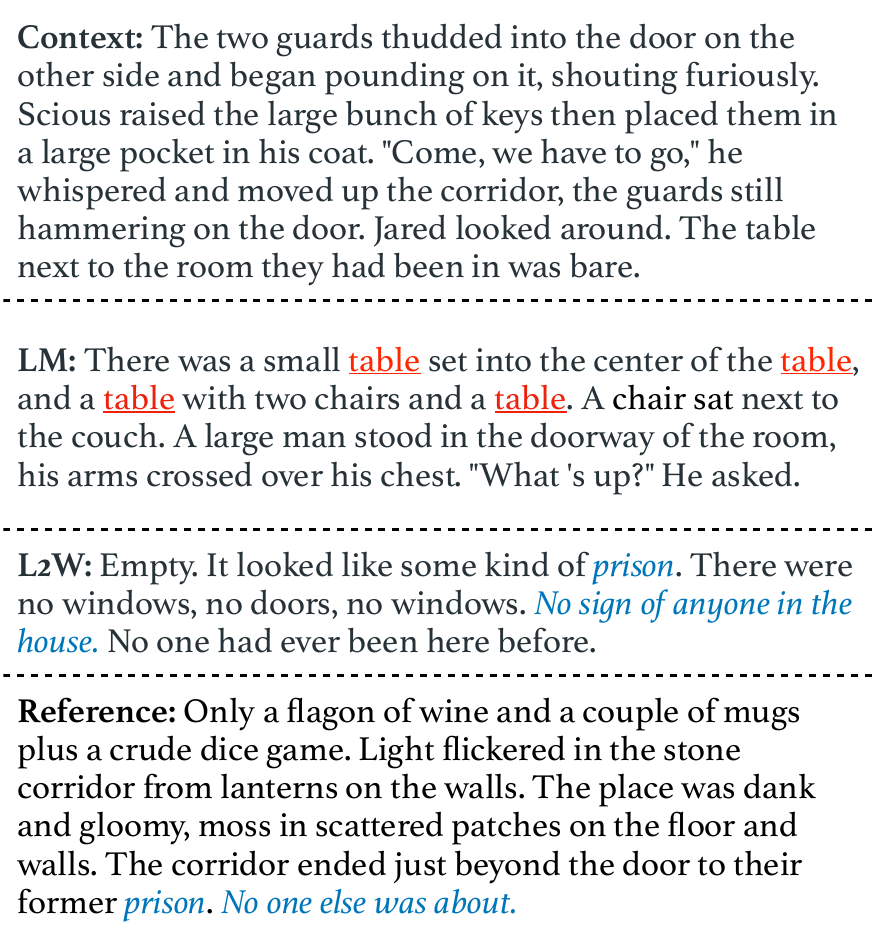}
\vspace*{-7mm}
\caption{Sample generations from an RNN language model (LM) and our system (L2W) conditioning on the context shown on the top. 
The {\color{myred}\underline{red, underlined}} text highlights 
repetitions, while the \textit{\color{myblue}blue, italicized} text highlights details that have a direct semantic parallel in the reference text.
}
\label{fig:opening}
\end{figure}

Language models based on Recurrent Neural Networks (RNNs) 
have 
brought substantial advancements across a wide range of language tasks \citep{jozefowicz2016exploring,bahdanau2014align,chopraAbs}.
However, when used for long-form text generation, RNNs often lead to degenerate text that is repetitive, self-contradictory, and overly generic, as shown in Figure~\ref{fig:opening}.

We propose a unified learning framework that can address several challenges of long-form text generation by composing a committee of discriminators each specializing in a different principle of communication. 
Starting with an RNN language model, our framework learns to construct a more powerful generator by training a number of discriminative models that can collectively address limitations of the base RNN generator, and then learns how to weigh these discriminators to form the final decoding objective. 
These ``cooperative'' discriminators complement each other and the base language model to form a stronger, more global decoding objective.

The design of our discriminators are inspired by Grice's maxims \citep{grice1975logic} of quantity, quality, relation, and manner.
The discriminators learn to encode these qualities through the selection of training data (e.g. distinguishing a true continuation from a randomly sampled one as in \S\hyperref[sssec:relevance]{3.2 Relevance Model}), which includes generations from partial models (e.g. distinguishing a true continuation from one generated by a language model as in  \S\hyperref[sssec:relevance]{3.2 Style Model}). 
The system then learns to balance these discriminators by initially weighing them uniformly, then continually updating its weights by comparing the scores the system gives to its own generated continuations and to the reference continuation.

Empirical results (\S\ref{sec:results}) demonstrate that our learning framework is highly effective in converting a generic RNN language model into a substantially stronger generator. Human evaluation confirms that language generated by our model is preferred over that of competitive baselines by a large margin in two distinct domains, and significantly enhances the overall coherence, style, and information content of the generated text. Automatic evaluation shows that our system is both less repetitive and more diverse than baselines.

\section{Background}

\label{sec:background}

RNN language models learn the conditional probability $P(x_t|x_{1},...,x_{t-1})$ of generating the next word $x_t$ given all previous words. 
This conditional probability learned by RNNs 
often assigns higher probability 
to repetitive, overly generic sentences, as shown in Figure~\ref{fig:opening} and also in Table~\ref{table:examples}. 
Even gated RNNs such as LSTMs \cite{hochreiter1997long} and GRUs \citep{cho2014properties} have 
difficulties in properly incorporating long-term context due to 
explaining-away effects \citep{yunoisy}, diminishing gradients  \citep{pascanu2013difficulty}, and lack of inductive bias for the network to learn discourse structure or global coherence beyond local patterns.  

Several methods in the literature attempt to address these issues. Overly simple and generic generation can be improved by length-normalizing the sentence probability \citep{Wu2016google}, future cost estimation \citep{Schmaltz2016ordering}, or a diversity-boosting objective function 
\citep{shao2017generating,vijayakumar2016diverse}. 
Repetition can be reduced by prohibiting recurrence of the trigrams as a hard rule \citep{Paulus2017reinforced}. 
However, such hard constraints do not stop RNNs from repeating through paraphrasing while preventing occasional intentional repetition. 

We propose a unified framework to address all these related challenges of long-form text generation by learning to construct a better decoding objective, generalizing over various existing modifications to the decoding objective. 


\section{The Learning Framework}
\label{sec:model}

We propose a general learning framework for conditional language generation of a sequence $\mathbf{y}$ given a fixed context $\mathbf{x}$. 
The decoding objective for generation takes the general form 
\begin{equation} 
f_{\lambda}(\mathbf{x},\mathbf{y}) = \log(P_{\textrm{\scriptsize lm}}(\mathbf{y} | \mathbf{x})) + \sum_k \lambda_k s_k(\mathbf{x}, \mathbf{y}), \label{eq:decoder-objective}
\end{equation}
where every $s_k$ is a scoring function.
The proposed objective combines the RNN language model probability $P_{\textrm{\scriptsize lm}}$ (\S\ref{ssec:base}) with a set of additional scores $s_k(\mathbf{x}, \mathbf{y})$ produced by discriminatively trained communication models (\S\ref{ssec:communication}), which are weighted with learned mixture coefficients $\lambda_k$ (\S\ref{ssec:metaweights}).
When the scores $s_k$ are log probabilities, this corresponds to a Product of Experts (PoE) model \citep{hinton2006poe}.

Generation is performed using beam search (\S\ref{ssec:beamsearch}),
scoring incomplete candidate generations $\mathbf{y}_{1:i}$
at each time step $i$. 
The RNN language model decomposes into per-word probabilities via the chain rule. 
However, in order to allow for more expressivity over long range context we do not require the discriminative model scores to factorize over the elements of $\mathbf{y}$, addressing a key limitation of RNNs. 
More specifically, we use an estimated score $s'_k(\mathbf{x}, \mathbf{y}_{1:i})$ that can be computed for any prefix of $\mathbf{y} = \mathbf{y}_{1:n}$ to approximate the objective during beam search, such that $s'_k(\mathbf{x}, \mathbf{y}_{1:n}) = s_k(\mathbf{x}, \mathbf{y})$. 
To ensure that the training method matches this approximation as closely as possible, scorers are trained to discriminate prefixes of the same length (chosen from a predetermined set of prefix lengths), rather than complete continuations, except for the entailment module as described in \S\hyperref[sssec:entailment]{3.2 Entailment Model}. The prefix scores are re-estimated at each time-step, rather than accumulated over beam search. 


\subsection{Base Language Model}
\label{ssec:base}

The RNN language model treats the context $\mathbf{x}$ and the continuation $\mathbf{y}$ as a single sequence $\mathbf{s}$: 
\begin{equation}
\log P_{\textrm{\scriptsize lm}}(\mathbf{s}) = \sum_i \log P_{\textrm{\scriptsize lm}}(\mathbf{s}_i | \mathbf{s}_{1:i-1}). 
\end{equation}

\subsection{Cooperative Communication Models} 
\label{ssec:communication}

We introduce a set of discriminators, each of which encodes an aspect of proper writing that RNNs usually fail to capture.
Each model is trained to discriminate between good and bad generations; we vary the model parameterization and training examples to guide each model to focus on a different aspect of Grice's Maxims. 
The discriminator scores are interpreted as classification probabilities  (scaled with the logistic function where necessary) and interpolated in the objective function as log probabilities.

Let $D = \{(\mathbf{x_1}, \mathbf{y_1}), \ldots (\mathbf{x_n}, \mathbf{y_n})\}$ be the set of training examples for conditional generation.
$D_{\mathbf{x}}$ denote all contexts and $D_{\mathbf{y}}$ all
continuations.
The scoring functions are trained on prefixes of $\mathbf{y}$ to simulate their application to partial continuations at inference time. 

In all models the first layer embeds each word $w$ into a 300-dimensional vector $e(w)$ initialized with GloVe \citep{pennington2014glove} pretrained-embeddings.

\subsubsection*{Repetition Model}
\label{sssec:repetition}

This model addresses the maxim of Quantity by biasing the generator to avoid repetitions. 
The goal of the repetition discriminator is to learn to distinguish between 
RNN-generated and gold continuations by exploiting our empirical
observation that repetitions are more common in completions
generated by RNN language models.
However, we do not want to completely eliminate repetition, as words do recur in English. 

In order to model natural levels of repetition, a score $d_i$ is computed for each position in the continuation $\mathbf{y}$ based on pairwise cosine similarity between word embeddings within a fixed window of the previous $k$ words, where
\begin{equation}
d_i = \max_{j=i-k\ldots i-1}(\textrm{CosSim}(e(y_j), e(y_i))),
\end{equation}
such that $d_i = 1$ if $y_i$ is repeated in the window.

The score of the continuation is then defined as
\begin{equation}
s_{\textrm{\footnotesize rep}}(\mathbf{y}) = \sigma (\mathbf{w}_{r}^\top \textrm{RNN}_{\textrm{\footnotesize rep}}(\mathbf{d})),
\end{equation}
where $\textrm{RNN}_{\textrm{\footnotesize rep}}(\mathbf{d})$ is the final state of a unidirectional RNN ran over the similarity scores $\mathbf{d} = d_1 \ldots d_n$ and $\mathbf{w}_r$ is a learned vector.
The model is trained to maximize the ranking log likelihood
\begin{equation}
L_{\textrm{\footnotesize rep}} = 
  \sum_{\substack{(\mathbf{x}, \mathbf{y}_g) \in D,\\
        \mathbf{y}_s \sim \textrm{LM}(\mathbf{x})}} 
 \log \sigma(s_{\textrm{\footnotesize rep}}(\mathbf{y}_g) -  s_{\textrm{\footnotesize rep}}(\mathbf{y}_{s})),
 \label{eq:reploss}
\end{equation}
which corresponds to the probability of the gold ending $\mathbf{y}_g$ receiving a higher score than the ending sampled from the RNN language model.

\subsubsection*{Entailment Model}
\label{sssec:entailment}
Judging textual quality can be related to the natural language inference (NLI) task of recognizing textual entailment \citep{Dagan2005rte,Bowman2015snli}: we would like to guide the generator to neither contradict its own past generation (the maxim of Quality) nor state something that readily follows from the context (the maxim of Quantity). 
The latter case is driven by the RNNs habit of paraphrasing itself during generation.

We train a classifier that takes two sentences $a$ and $b$ as input and predicts the relation between them as either \emph{contradiction}, \emph{entailment} or \emph{neutral}. We use the \emph{neutral} class probability of the sentence pair as discriminator score, in order to discourage both contradiction and entailment.
As entailment classifier we use the decomposable attention model \cite{parikh2016decomposable}, a competitive, parameter-efficient model for entailment classification.\footnote{We use the version without intra-sentence attention.}
The classifier is trained on two large entailment datasets, SNLI \cite{Bowman2015snli} and MultiNLI \citep{Williams2017multisnli}, which together have more than 940,000 training examples.
We train separate models based on the vocabularies of each of the datasets we use for evaluation. 

In contrast to our other communication models, this classifier cannot be applied directly to the full context and continuation sequences it is scoring. 
Instead every completed sentence in the continuation should be scored against all preceding sentences in both the context and continuation. 

Let $t(\mathbf{a}, \mathbf{b})$ be the log probability of the neutral class. Let $S(\mathbf{y})$ be the set of complete sentences in $\mathbf{y}$, $S_{\textrm{last}}(\mathbf{y})$ the last complete sentence, and $S_{\textrm{init}}(\mathbf{y})$ the sentences before the last complete sentence.
We compute the entailment score of $S_{\textrm{last}}(\mathbf{y})$ against all preceding sentences in $\mathbf{x}$ and $\mathbf{y}$, and use the score of the sentence-pair for which we have the least confidence in a \emph{neutral} classification:
\begin{equation}
s_{\textrm{\footnotesize entail}}(\mathbf{x}, \mathbf{y}) = min_{\mathbf{a} \in S(\mathbf{x}) \cup S_{\textrm{init}}(\mathbf{y})} t(\mathbf{a}, S_{\textrm{last}}(\mathbf{y})).
\end{equation}

Intuitively, we only use complete sentences because the ending of a sentence can easily flip entailment. As a result, we carry over entailment score of the last complete sentence in a generation until the end of the next sentence, in order to maintain the presence of the entailment score in the objective.
Note that we check that the current sentence is not directly entailed or contradicted by a previous sentence and not the reverse. 
\footnote{If the current sentence entails a previous one it may simply be adding more specific information, for instance: ``He hated broccoli. Every time he ate broccoli he was reminded that it was the thing he hated most.''}
In contrast to our other models, the score this model returns only corresponds to a subsequence of the given continuation, as the score is not accumulated across sentences during beam search.
Instead the decoder is guided locally to continue complete sentences that are not entailed or contradicted by the previous text.

\begin{algorithm}[t]
    \small
 \KwData{context $\mathbf{x}$, beam size $k$, sampling temperature $t$}
 \KwResult{best continuation}
 best = None \\
 beam = [$\mathbf{x}$]\\ 
 \For{\upshape step = 0; step $<$ max\_steps; step = step $ + 1$}{
  next\_beam = [] \\
  \For{\upshape candidate in beam} {
    next\_beam.extend(next\_k(candidate)) \\
    \If{ \upshape termination\_score(candidate) $>$ best.score } {
        best = candidate.append(term)   
    }
  }
  \For{\upshape candidate in next\_beam} {
    \Comment{score with models} \\
    candidate.score += $f_{\lambda}(\textrm{candidate})$ 
  }
  \Comment{sample k candidates by score} \\
  beam = sample(next\_beam, $k$, $t$)
 }
 \If{\upshape learning} {
    update $\lambda$ with gradient descent by comparing best against the gold. 
 }
 return best
 \vspace*{4mm}
\caption{Inference/Learning in the Learning to Write Framework.
}
\label{fig:algo}
\end{algorithm}

\begin{table*}\small

\begin{center}
\begin{tabular}{c||cc|ccc|cc|ccc}
\hline
& \multicolumn{5}{c|}{\bf BookCorpus} & \multicolumn{5}{c}{\bf TripAdvisor} \\
\hline
\textbf{Model} & BLEU & Meteor  & Length & Vocab & Trigrams & BLEU & Meteor & Length & Vocab \% & Trigrams \\ 
\hline
\textsc{L2W}  &  \textbf{0.52} & \textbf{6.8} & \textbf{43.6} & \textbf{73.8} & 98.9  & 1.7 & 11.0 & 83.8 & \textbf{64.1} & \textbf{96.2} \\
\hline
\textsc{AdaptiveLM} & \textbf{0.52} &  6.3 & 43.5 & 59.0 & 92.7  & \textbf{1.94} & \textbf{11.2} & \textbf{94.1} & 52.6 & 92.5 \\
\textsc{CacheLM} & 0.33 & 4.6 & 37.9 & 31.0  & 44.9  & 1.36 & 7.2 & 52.1 & 39.2 & 57.0 \\
\textsc{Seq2Seq} & 0.32 & 4.0 & 36.7 & 23.0 & 33.7  & 1.84 & 8.0 & 59.2 & 33.9 & 57.0 \\
\textsc{SeqGAN} & 0.18 & 5.0 & 28.4 & 73.4 & \textbf{99.3} & 0.73 & 6.7 & 47.0 & 57.6 & 93.4 \\
\hline
\textsc{Reference} & 100.0 & 100.0 & 65.9 & 73.3 & 99.7  & 100.0 & 100.0 & 92.8 & 69.4 & 99.4 \\
\hline
\end{tabular}
\end{center}
\caption{Results for automatic evaluation metrics for all systems and domains, using the original continuation as the reference. The metrics are: Length - Average total length per example; Trigrams - \% unique trigrams per example; Vocab - \% unique words per example.
}

\label{table:autotable1}

\end{table*}

\subsubsection*{Relevance Model}
\label{sssec:relevance}

The relevance model encodes the maxim of Relation by predicting whether the content of a candidate continuation is relevant to the given context.
We train the model to distinguish between true continuations and random continuations sampled from other (human-written) endings in the corpus, conditioned on the given context.

First both the context and continuation sequences are passed through a convolutional layer, followed by maxpooling to obtain vector representations of the sequences:
\begin{align}
a &= \textrm{maxpool}(\textrm{conv}_a (e(\mathbf{x}))), \\
b &= \textrm{maxpool}(\textrm{conv}_b (e(\mathbf{y}))).
\end{align}
The goal of maxpooling is to obtain a vector representing the most important semantic information in each dimension.

The scoring function is then defined as 
\begin{equation}
s_{\textrm{\footnotesize rel}} = \mathbf{w}_{l}^T \cdot(a \circ b),
\end{equation}
where element-wise multiplication of the context and continuation vectors will amplify similarities. 

We optimize the ranking log likelihood
\begin{equation}
L_{\textrm{\footnotesize rel}} =   
  \sum_{\substack{(\mathbf{x}, \mathbf{y}_g) \in D, \\ 
         \mathbf{y}_r \sim D_{\mathbf{y}}}} 
  \log \sigma(s_{\textrm{\footnotesize rel}}(\mathbf{x}, \mathbf{y}_g) -  s_{\textrm{\footnotesize rel}}(\mathbf{x}, \mathbf{y}_{r})),    
\end{equation}
where $\mathbf{y}_g$ is the gold ending and $\mathbf{y}_r$ is a randomly sampled ending.

\subsubsection*{Lexical Style Model}
\label{sssec:style}

In practice RNNs generate text that exhibit much less lexical diversity than their training data. 
To counter this effect we introduce a simple discriminator based on observed lexical distributions which captures writing style as expressed through word choice. 
This classifier therefore encodes aspects of the maxim of Manner.

The scoring function is defined as
\begin{equation}
s_{\textrm{\footnotesize bow}}(\mathbf{y}) = \mathbf{w}_{s}^T \textrm{maxpool}(e(\mathbf{y})).
\end{equation}

The model is trained with a ranking loss using negative examples sampled from the language model, similar to Equation \ref{eq:reploss}.

\subsection{Mixture Weight Learning}
\label{ssec:metaweights}


Once all the communication models have been trained, we learn the combined decoding objective.
In particular we learn the weight coefficients $\lambda_k$ in equation \ref{eq:decoder-objective} to linearly combine the scoring functions, using a discriminative loss
\begin{equation}
L_{\textrm{\footnotesize mix}} = \sum_{(\mathbf{x}, \mathbf{y}) \in D}   (f_{\lambda}(\mathbf{x}, \mathbf{y}) - f_{\lambda}(\mathbf{x}, \mathcal{A}(\mathbf{x}))^2, 
\end{equation}
where $\mathcal{A}$ is the inference algorithm for beam search
decoding. 
The weight coefficients are thus optimized to minimize the difference between the scores assigned to the gold continuation and the continuation predicted by the current model.

Mixture weights are learned online: Each successive generation is performed
based on the current values of $\lambda$, and a step of gradient
descent is then performed based on the prediction.
This has the effect that the objective function changes dynamically during training: 
As the current samples from the model are used to update the mixture weights, it creates its own learning signal by applying the generative model discriminatively. 
The SGD learning rate is tuned separately for each dataset.

\subsection{Beam Search}
\label{ssec:beamsearch}

Due to the limitations of greedy decoding and the fact that our scoring functions do not decompose across time steps, we perform generation with a beam search procedure, shown in Algorithm~\ref{fig:algo}.
The naive approach would be to perform beam search based only 
on the language model, and then rescore the $k$ best candidate
completions with our full model.
We found that this approach leads to limited diversity in the beam
and therefore cannot exploit the strengths of the full model.

Instead we score the current hypotheses in the beam with the full decoding objective:
First, each hypothesis is expanded by selecting the $k$ highest scoring next words according to the language model (we use beam size $k=10$). 
Then $k$ sequences are sampled from the $k^2$ candidates
according to the (softmax normalized) distribution over the candidate scores given by the full decoding objective. 
Sampling is performed in order to increase diversity, using
a temperature of $1.8$, which was tuned by comparing the coherence of continuations on the validation set.

At each step, the discriminator scores are recomputed for all candidates, with the exception of the entailment score, which is only recomputed for hypotheses which end with a sentence terminating symbol.
We terminate beam search when the \emph{termination\_score}, the maximum possible score achievable by terminating generation at the current position, is smaller than the current best score. 

\begin{table*}\small

\begin{center}
\begin{tabular}{ c|cccc||ccc }
\hline
\textbf{BookCorpus}
& \multicolumn{4}{c||}{\bf Specific Criteria} & \multicolumn{3}{c}{\bf Overall Quality} \\
\hline
\textbf{L2W vs.} & Repetition & Contradiction & Relevance  & Clarity &  Better & Equal &  Worse\\
\hline
\textsc{AdaptiveLM} & +0.48 & +0.18 & +0.12 & +0.11 & \textbf{47\%} & 20\% & 32\% \\
\textsc{CacheLM} & +1.61 & +0.37 & +1.23 & +1.21 & \textbf{86\%} & 6\% & 8\%  \\
\textsc{Seq2Seq} & +1.01 & +0.54 & +0.83 & +0.83 & \textbf{72\%} & 7\% & 21\% \\
\textsc{SeqGAN} & +0.20 & +0.32 & +0.61 & +0.62 & \textbf{63\%} & 20\%& 17\% \\
\hline
\textsc{LM vs. Reference} & -0.10 & -0.07 & -0.18 & -0.10 & 41\% & 7 \% & \textbf{52}\% \\
\textsc{L2W vs. Reference} & +0.49 & +0.37 & +0.46 & +0.55 & \textbf{53\%} & 18\% & 29\% \\
\hline
\end{tabular}
\end{center}

\begin{center}
\begin{tabular}{ c|cccc||ccc}
\hline
\textbf{TripAdvisor}
 & \multicolumn{4}{c||}{\bf Specific Criteria} & \multicolumn{3}{c}{\bf Overall Quality} \\
 \hline
\textbf{L2W vs.} & Repetition & Contradiction & Relevance  & Clarity &  Better & Equal &  Worse\\
\hline
\textsc{AdaptiveLM} & +0.23 & -0.02 & +0.19 & -0.03 & \textbf{47\%} & 19\% & 34\%   \\
\textsc{CacheLM} & +1.25 & +0.12 & +0.94 & +0.69 & \textbf{77\%} & 9\% & 14\%  \\
\textsc{Seq2Seq} & +0.64 & +0.04 & +0.50 & +0.41 &  \textbf{58\%} & 12\% & 30\% \\
\textsc{SeqGAN} & +0.53 & +0.01 & +0.49 & +0.06 & \textbf{55\%} & 22\% & 22\% \\
\hline
\textsc{LM vs. Reference} & -0.10 & -0.04 & -0.15 & -0.06 & 38\% & 10\% & \textbf{52\%} \\
\textsc{L2W vs. Reference} & -0.49 & -0.36 & -0.47 & -0.50 & 25\% & 18\% & \textbf{57\%} \\
\hline
\end{tabular}
\end{center}

\caption{Results of crowd-sourced evaluation on different aspects of the generation quality as well as overall quality judgments. For each sub-criteria we report the average of comparative scores on a scale from -2 to 2. For the overall quality evaluation decisions are aggregated over 3 annotators per example.
}

\label{table:bigtable}

\end{table*}

\section{Experiments}
\label{sec:experiments}

\subsection{Corpora}

We use two English corpora for evaluation. 
The first is the TripAdvisor corpus \citep{wang2010latent}, a collection of hotel reviews with a total of 330 million words.\footnote{\url{http://times.cs.uiuc.edu/~wang296/Data/}} 
The second is the BookCorpus \citep{moviebook}, a 980 million word collection of novels 
by unpublished authors.\footnote{\url{http://yknzhu.wixsite.com/mbweb}} In order to train the discriminators, mixing weights, and the \textsc{Seq2Seq} and \textsc{SeqGAN} baselines, we segment both corpora into sections of length ten sentences, and use the first 5 sentence as context and the second 5 as the continuation. See Appendix~\ref{app:corp} for further details.

\subsection{Baselines}

\paragraph{\textsc{AdaptiveLM}}
Our first baseline is the same Adaptive Softmax \citep{grave2016efficient} language model used as base generator in our framework (\S\ref{ssec:base}).
This enables us to evaluate the effect of our enhanced decoding objective directly. A $100$k vocabulary is used and beam search with beam size of $5$ is used at decoding time. \textsc{AdaptiveLM} achieves perplexity of 37.46 and 18.81 on BookCorpus and TripAdvisor respectively. 
\vspace{-5.0pt}
\paragraph{\textsc{CacheLM}}
As another LM baseline we include a continuous cache language model \citep{grave2016cache} as implemented by \citet{merityRegOpt}, which recently obtained state-of-the-art perplexity on the Penn Treebank corpus \citep{marcus1993building}. Due to memory constraints, we use a vocabulary size of $50$k for \textsc{CacheLM}. To generate, beam search decoding is used with a beam size $5$. \textsc{CacheLM} obtains perplexities of $70.9$ and $29.71$ on BookCorpus and TripAdvisor respectively.
\vspace{-5.0pt}
\paragraph{\textsc{Seq2Seq}}
As our evaluation can be framed as sequence-to-sequence transduction, we compare against a seq2seq model directly trained to predict 5 sentence continuations from 5 sentences of context, using the OpenNMT attention-based seq2seq implementation \citep{opennmt}. Similarly to \textsc{CacheLM}, a $50$k vocabulary was used and beam search decoding was performed with a beam size of $5$.
\vspace{-5.0pt}
\paragraph{\textsc{SeqGAN}}
Finally, as our use of discriminators is related to Generative Adversarial Networks (GANs), we use SeqGAN \citep{yu2017seqgan}, a GAN for discrete sequences trained with policy gradients.\footnote{We use the implementation available at \url{https://github.com/nhynes/abc}. }
This model is trained on 10 sentence sequences, which is significantly longer than previous experiments with GANs for text; the vocabulary is restricted to $25$k words to make training tractable. Greedy sampling was found to outperform beam search. For implementation details see Appendix~\ref{app:base}.
\begin{table*} \small
\begin{tabular}{ c|p{5.2in} }
\hline
\hline
\sc Context & We thoroughly enjoyed staying at this hotel. We travelled with our kids aged 4 and 6. The rooms were clean, spacious and well maintained. The beds were great and all of us slept well. The hotel provided vouchers towards breakfast at the adjoining restaurant. \\
\hline
\sc L2W & We didn't have breakfast, so can't comment on that. The concierge was very helpful, and gave us directions, etc. Overall, we would definitely recommend this hotel to others. We would stay here again in a heartbeat. If you're looking for a great location, and good value, I'd recommend this hotel.\\
\hline
\sc AdaptiveLM &  The breakfast buffet was very good, and the staff was very friendly and helpful, and it was a great way to start our day! I would definitely stay here again, and would recommend it to my friends, and would stay there again if I ever go back to Montreal. The location was great, and the staff was friendly and helpful. I would definitely stay here again!! \\ 
\hline
\sc CacheLM &  The staff was very friendly and helpful. The hotel is located in a great location. It is right in the middle of the gaslamp district.     I would recommend this hotel to anyone. I would definitely stay here again. \\ 
\hline
\sc Seq2Seq &   The staff were very friendly and helpful. The hotel is in a great location. The staff were very friendly and helpful. The hotel is in a great location. The staff were very friendly and helpful. \\ 
\hline
\sc SeqGAN &   We had a breakfast at Shula's \& a delicious breakfast. The staff was very helpful and helpful. The breakfast was great as well. The staff was very helpful and friendly. We had a great service and the food was excellent. \\ 
\hline
\sc Reference &  The restaurant was great and we used the vouchers towards whatever breakfast we ordered. The hotel had amazing grounds with a putting golf course that was fun for everyone. The pool was fantastic and we lucked out with great weather. We spent many hours in the pool, lounging, playing shuffleboard and snacking from the attached bar. The happy hour was great perk. \\
\end{tabular}
\caption{Example continuations generated by our model (L2W) and various baselines (all given the same context from TripAdvisor) compared to the reference continuation. For more examples go to \url{https://ari-holtzman.github.io/l2w-demo/}.
}

\label{table:examples}
\end{table*}

\subsection{Evaluation Setup}

We pose the evaluation of our model as the task of generating an appropriate continuation given an initial context.
In our open-ended generation setting the continuation is not required to be a specific length, so we require our models and baselines to generate $5$-sentence continuations, consistent with the way the discriminator and seq2seq baseline datasets are constructed.

Previous work has reported that automatic measures such as 
BLEU \citep{papineni2002bleu} and Meteor \citep{denkowski2010meteor} do not lead to meaningful evaluation when used for long or creative text generation where there can be high variance among acceptable generation outputs \citep{wiseman2017challenges,cider}. 
However, we still report these measures as one component of our evaluation. 
Additionally we report a number of custom metrics which capture important properties of the generated text:
\emph{Length} -- Average sequence length per example; \emph{Trigrams} -- percentage of unique trigrams per example; \emph{Vocab} -- percentage of unique words per example.
Endings generated by our model and the baselines are compared against the reference endings in the original text. 
Results are given in Table \ref{table:autotable1}.

For open-ended generation tasks such as our own, human evaluation has been found to be the only reliable measure \citep{JiweiRobotChat,wiseman2017challenges}. 
For human evaluation, two possible endings are presented to a human, who assesses the text according to several criteria, 
which are closely inspired by Grice's Maxims: repetition, contradiction, relevance and clarity. See Appendix~\ref{app:eval} for examples of the evaluation forms we used.
For each criterion, the two continuations are compared using a 5-point Likert scale, to which we assign numerical values of $-2$ to $2$.
The scale measures whether one generation is strongly or somewhat preferred above the other, or whether they are equal.
Finally, the human is asked to make a judgement about overall quality: which ending is better, or are they of equal quality?

The human evaluation is performed on $100$ examples selected from the test set of each corpus, for every pair of generators that are compared.
We present the examples to workers on Amazon Mechanical Turk, using three annotators for each example. 
The results are given in Table~\ref{table:bigtable}. 
For the Likert scale, we report the average scores for each criterion, while for the overall quality judgement we simply aggregate votes across all examples.

\section{Results and Analysis}
\label{sec:results}

\subsection{Quantitative Results}

\begin{table*}\small

\textbf{Trip Advisor Ablation}
\begin{center}
\begin{tabular}{ c|cccc||ccc}
\hline
\textbf{Ablation vs. LM} & Repetition & Contradiction & Relevance  & Clarity &  Better & Neither &  Worse\\
\hline
\textsc{Repetition Only} &  +0.63 & +0.30 & +0.37 & +0.42 & \textbf{50}\% & 23\% & 27\%  \\
\textsc{Entailment Only} & +0.01 & +0.02 & +0.05 & -0.10  & 39\%   & 20\% & \textbf{41\%}  \\
\textsc{Relevance Only} & -0.19 & +0.09 & +0.10 & +0.060 & 36\% & 22\% & \textbf{42\%} \\
\textsc{Lexical Style Only} & +0.11 & +0.16 & +0.20 & +0.16 & 38\% & 25\% &  \textbf{38}\% \\
\hline
\textsc{All} & +0.23 & -0.02 & +0.19 & -0.03 & \textbf{47\%} & 19\% & 34\%   \\
\hline
\end{tabular}
\end{center}

\caption{Crowd-sourced ablation evaluation of generations on TripAdvisor. Each ablation uses only one discriminative communication model, and is compared to \textsc{AdaptiveLM}.
}
\label{table:ablationtable}

\end{table*}

The absolute performance of all the evaluated systems on BLEU and Meteor is quite low (Table \ref{table:autotable1}), as expected. 
However, in relative terms \textsc{L2W} is superior or competitive with all the baselines, of which \textsc{AdaptiveLM} performs best. In terms of vocabulary and trigram diversity only \textsc{SeqGAN} is competitive with \textsc{L2W}, likely due to the fact that sampling based decoding was used. For generation length only \textsc{L2W} and \textsc{AdaptiveLM} even approach human levels, with the former better on BookCorpus and the latter on TripAdvisor.

Under the crowd-sourced evaluation (Table \ref{table:bigtable}),  
on BookCorpus our model is consistently favored over the baselines on all dimensions of comparison. In particular, our model tends to be much less repetitive, while being more clear and relevant than the baselines.
\textsc{AdaptiveLM} is the most competitive baseline owing partially to the robustness of language models and to greater vocabulary coverage through the adaptive softmax. \textsc{SeqGAN}, while failing to achieve strong coherency, is surprisingly diverse, but tended to produce far shorter sentences than the other models. \textsc{CacheLM} has trouble dealing with the complex vocabulary of our domains without the support of either a hierarchical vocabulary structure (as in \textsc{AdaptiveLM}) or a structured training method (as with \textsc{SeqGAN}), leading to overall poor results.
While the \textsc{Seq2Seq} model has low conditional perplexity, we found that in practice it is less able to leverage long-distance dependencies than the base language model, producing more generic output. This reflects our need for more complex evaluations for generation, as such models are rarely evaluated under metrics that inspect \textit{characteristics} of the text, rather than ability to predict the gold or overlap with the gold. 

For the TripAdvisor corpus, \textsc{L2W} is ranked higher than the baselines on overall quality, as well as on most individual metrics, with the exception that it fails to improve on contradiction and clarity over the \textsc{AdaptiveLM} (which is again the most competitive baseline). 
Our model's strongest improvements over the baselines are on repetition and relevance. 

\subsubsection*{Ablation}

To investigate the effect of individual discriminators on the overall performance,  we report the results of ablations of our model in Table~\ref{table:ablationtable}.
For each ablation we include only one of the communication modules, and train a single mixture coefficient for combining that module and the language model. 
The diagonal of Table~\ref{table:ablationtable} contains only positive numbers, indicating that each discriminator does help with the purpose it was designed for. Interestingly, most discriminators help with most aspects of writing, but all except repetition fail to actually improve the overall quality over \textsc{AdaptiveLM}.

The repetition module gives the largest boost by far, consistent with the intuition that many of the deficiencies of RNN as a text generator lie in semantic repetition. The entailment module (which was intended to reduce contradiction) is the weakest, which we hypothesize is the combination of (a) mismatch between training and test data (since the entailment module was trained on SNLI and MultiNLI) and (b) the lack of smoothness in the entailment scorer, whose score could only be updated upon the completion of a sentence.

\subsubsection*{Crowd Sourcing}
Surprisingly, \textsc{L2W} is even preferred over the \textit{original} continuation of the initial text on BookCorpus. Qualitative analysis shows that \textsc{L2W}'s continuation is often a straightforward continuation of the original text while the true continuation is more surprising and contains complex references to earlier parts of the book. While many of the issues of automatic metrics \citep{liuhownot,novikovaWhy} have been alleviated by crowd-sourcing, we found it difficult to incentivize crowd workers to spend significant time on any one datum, forcing them to rely on a shallower understanding of the text.

\subsection{Qualitative Analysis}
\label{ssec:qual}

\textsc{L2W} generations are more topical and stylistically coherent with the context than the baselines. Table~\ref{table:examples} shows that \textsc{L2W}, \textsc{AdaptiveLM}, and \textsc{SeqGAN} all start similarly, commenting on the breakfast buffet, as breakfast was mentioned in the last sentence of the context. The language model immediately offers generic compliments about the breakfast and staff, whereas \textsc{L2W} chooses a reasonable but less obvious path, stating that the previously mentioned vouchers were not used. In fact, \textsc{L2W} is the only system not to use the line \emph{``The staff was very friendly and helpful.''}, despite this sentence appearing in less than 1\% of reviews.
The semantics of this sentence, however, is expressed in many different surface forms in the training data (e.g., \textit{``The staff were kind and quick to respond.''}).

The \textsc{CacheLM} begins by generating the same over-used sentence 
and only produce short, generic sentences throughout. Seq2Seq simply repeats sentences that occur often in the training set, repeating one sentence three times and another twice. 
This indicates that the encoded context is essentially being ignored
 as the model fails to align the context and continuation. 

The \textsc{SeqGAN} system is more detailed, e.g. mentioning a specific location ``Shula's'' as would be expected given its highly diverse vocabulary (as seen in Table~\ref{table:autotable1}). Yet it repeats itself in the first sentence. (e.g. \emph{``had a breakfast'', ``and a delicious breakfast''}). Consequently \textsc{SeqGAN} quickly devolves into generic language, repeating the incredibly common sentence \emph{``The staff was very helpful and friendly.''}, similar to \textsc{Seq2Seq}. 

The \textsc{L2W} models do not fix every degenerate characteristic of RNNs. The TripAdvisor \textsc{L2W} generation consists of meaningful but mostly disconnected sentences, whereas human text tends to build on previous sentences, as in the reference continuation. Furthermore, while \textsc{L2W} repeats itself less than any of our baselines, it still paraphrases itself, albeit more subtly: ``we would definitely recommend this hotel to others.'' compared to ``I'd recommend this hotel.'' 
This example also exposes a more fine-grained issue: \textsc{L2W} switches from using ``we'' to using ``I'' mid-generation. Such subtle distinctions are hard to capture during beam re-ranking and none of our models address the linguistic issues of this subtlety.

\section{Related Work}
\label{sec:related}

\paragraph{Alternative Decoding Objectives}

A number of papers have proposed \emph{alternative decoding objectives} for generation \citep{shao2017generating}. 
\citet{Li2016diversity} proposed a \emph{diversity-promoting objective} that interpolates the conditional probability score with negative marginal or reverse conditional probabilities. 
 \citet{yunoisy} also incorporate the reverse conditional probability through a noisy channel model in order to alleviate the \emph{explaining-away} problem, but at the cost of significant decoding complexity, making it impractical for paragraph generation. 
Modified decoding objectives have long been a common practice in statistical machine translation \citep{Koehn2003phrase,och2003mert,Watanabe2007mira,chiang2009features} and remain common with neural machine translation, even when an extremely large amount of data is available \citep{Wu2016google}. 
Inspired by all the above approaches, our work presents a general learning framework together with a more comprehensive set of composite communication models. 

\paragraph{Pragmatic Communication Models}
Models for pragmatic reasoning about communicative goals such as Grice's maxims have been proposed in the context of referring expression generation \citep{Frank2012pragmatic}. 
\citet{Andreas2016pragmatics} proposed a neural model where candidate descriptions are sampled from a generatively trained \emph{speaker}, which are then re-ranked by interpolating the score with that of the \emph{listener}, a discriminator that predicts a distribution over choices given the speaker's description.
Similar to our work the generator and discriminator scores are combined to select utterances which follow Grice's maxims.
\citet{yu2017joint} proposed a model where the speaker consists of a convolutional encoder and an LSTM decoder, trained with a ranking loss on negative samples in addition to optimizing log-likelihood.

\paragraph{Generative Adversarial Networks}
GANs \citep{goodfellow2014gans} are another alternative to maximum likelihood estimation for generative models. 
However, backpropagating through discrete sequences and the inherent instability of the training objective \citep{Che2017gans} both present significant challenges.
While solutions have been proposed to make it possible to train GANs for 
language \citep{Che2017gans,yu2017seqgan} they have not yet been shown to produce high quality long-form text, as our results confirm.

\paragraph{Generation with Long-term Context}

Several prior works studied paragraph generation using sequence-to-sequence models for image captions \citep{krause2016paragraphs}, product reviews \citep{Lipton2015reviews,dong2017reviews}, sport reports \citep{wiseman2017challenges}, and recipes \citep{kiddon2016checklist}. 
%
While these prior works focus on developing neural architectures for learning domain specific discourse patterns, our work proposes a general framework for learning a generator that is more powerful than maximum likelihood decoding from an RNN language model for an arbitrary target domain.
%




\section{Conclusion}
\label{sec:conclusion}

We proposed a unified learning framework for the generation of long, coherent texts, which overcomes some of the common limitations of RNNs as text generation models. Our framework learns a decoding objective suitable for generation through a learned combination of sub-models that capture linguistically-motivated qualities of good writing. Human evaluation shows that the quality of the text produced by our model exceeds that of competitive baselines by a large margin.

\section*{Acknowledgments}

We  thank the anonymous reviewers for their insightful feedback and Omer Levy for helpful discussions. 
This research was supported in part by NSF (IIS-1524371), DARPA CwC  through ARO (W911NF-15-1-0543), Samsung AI Research, and gifts by Tencent, Google, and Facebook.

\bibliography{conference}
\bibliographystyle{acl_natbib}

\appendix

\section[A]{Model Details}
\label{app:mod}

\subsection[A]{Base Language Model}

We use a 2-layer GRU \cite{cho2014properties} with a hidden size of 1024 for each layer.
Following \cite{inan2017tying} we tie the input and output embedding layers' parameters. We use an Adaptive Softmax for the final layer \citep{grave2016efficient}, which factorizes the prediction of a token into first predicting the probability of $k$ (in our case $k=3$) clusters of words that partition the vocabulary and then the probability of each word in a given cluster.
To regularize we dropout \citep{srivastava2014dropout} cells in the output layer of the first layer with probability 0.2. 
We use mini-batch stochastic gradient descent (SGD) and anneal the learning rate when the validation set performance fails to improve, checking every 1000 batches. Learning rate, annealing rate, and batch size were tuned on the validation set for each dataset.
Gradients are backpropagated 35 time steps and clipped to a maximum value of 0.25. 

\subsection{Cooperative Communication Models}

For all the models except the entailment model, training is performed with Adam \citep{kingma2014adam} with batch size $64$ and learning rate $0.01$.
The classifier's hidden layer size is $300$.
Dropout is performed on both the input word embeddings and the  non-linear hidden layer before classification with rate $0.5$. 

Word embeddings are kept fixed during training for the repetition model, but are fine-tuned for all the other models. 

\subsubsection*{Entailment Model}

We mostly follow the hyperparameters of \citet{parikh2016decomposable}:
Word embeddings are projected to a hidden size of $200$, which are used throughout the model. 
Optimization is performed with AdaGrad \cite{duchi2011adaptive} with initial learning rate $1.0$ and batch size $16$. 
Dropout is performed at rate $0.2$ on the hidden layers of the  2-layer MLPs in the model.

Our entailment classifier obtains $82\%$ accuracy on the SNLI validation set  and $68\%$ accuracy on the MultiNLI validation set.
\subsubsection*{Relevance Model}

The convolutional layer is a one-dimensional convolution with filter size 3 and stride 1; the input sequences are padded such that the input and output lengths are equal.

\section[B]{Baseline Details}
\label{app:base}

\paragraph{\textsc{CacheLM}}
Due to memory constraints, we use a vocabulary size of $50$k for \textsc{CacheLM}. Beam search decoding is used, with a beam size $5$.

\paragraph{\textsc{SeqGAN}}

The implementation we used adds a number of modelling extensions to the original SeqGAN.
In order to make training tractable, the vocabulary is restricted to $25$k words, the maximum sequence length is restricted to $250$, Monte Carlo rollouts to length 4, and the discriminator updated once for every $10$ generator training steps.
Greedy decoding sampling with temperature $0.7$ was found to work better than beam search.

\paragraph{\textsc{Seq2Seq}}
Due to memory constraints, we use a vocabulary size of $50$k for \textsc{Seq2Seq}. Beam search decoding is used, with a beam size $5$.

\section[C]{Corpora}
\label{app:corp}

For the language model and discriminators we use a vocabulary of $100,000$ words -- we found empirically that larger vocabularies lead to better generation quality. 
To train our discriminators and evaluate our models, we use segments of length 10, using the first 5 sentences as context and the second 5 as the reference continuation.
For TripAdvisor we use the first 10 sentences of reviews of length at least 10. 
For the BookCorpus we split books into segments of length 10.
We select $20\%$ of each corpus as held-out data (the rest is used for language model training). 
From the held-out data we select a test set of $2000$ examples and two validation sets of $1000$ examples each, one of which is used to train the mixture weights of the decoding objective.  
The rest of the held-out data is used to train the discriminative classifiers.

\section[D]{Evaluation Setup}
\label{app:eval}

The forms used on Amazon Mechanical Turk are pictured in Tables \ref{table:tbooks1}, \ref{table:tbooks2}, \ref{table:trip1}, and \ref{table:trip2}.

\begin{table*}
\center{\textbf{BookCorpus}}
\includegraphics[width=6in]{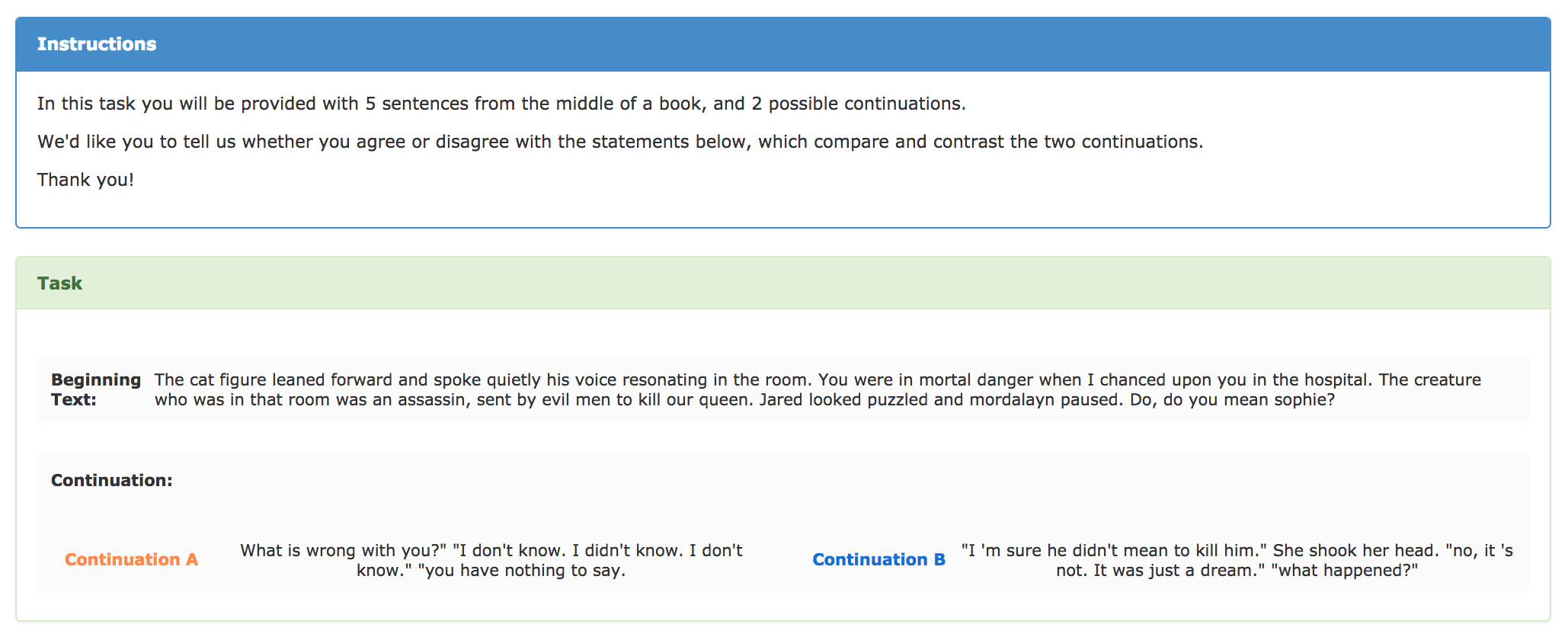}
\includegraphics[width=6in]{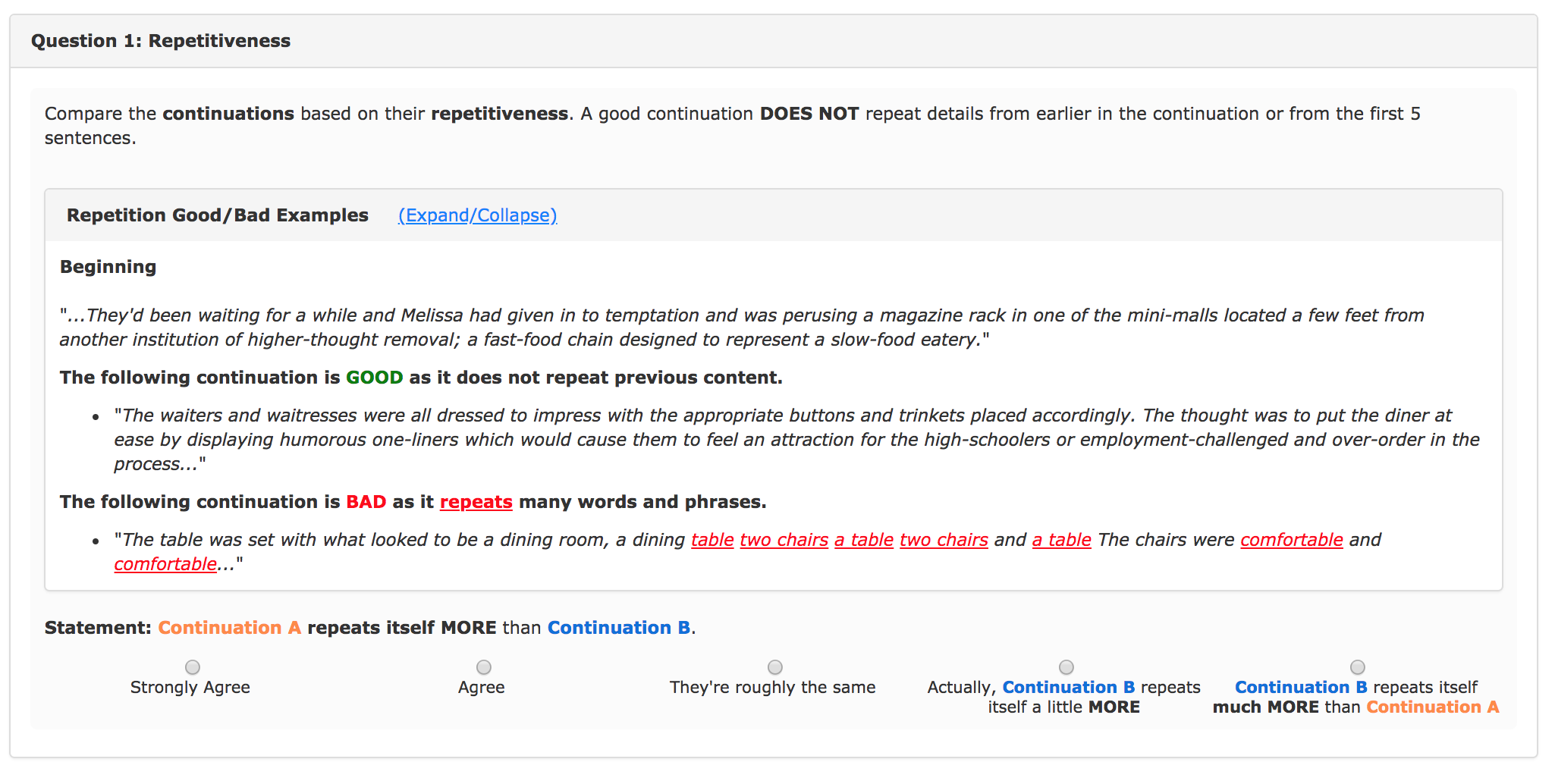}
\includegraphics[width=6in]{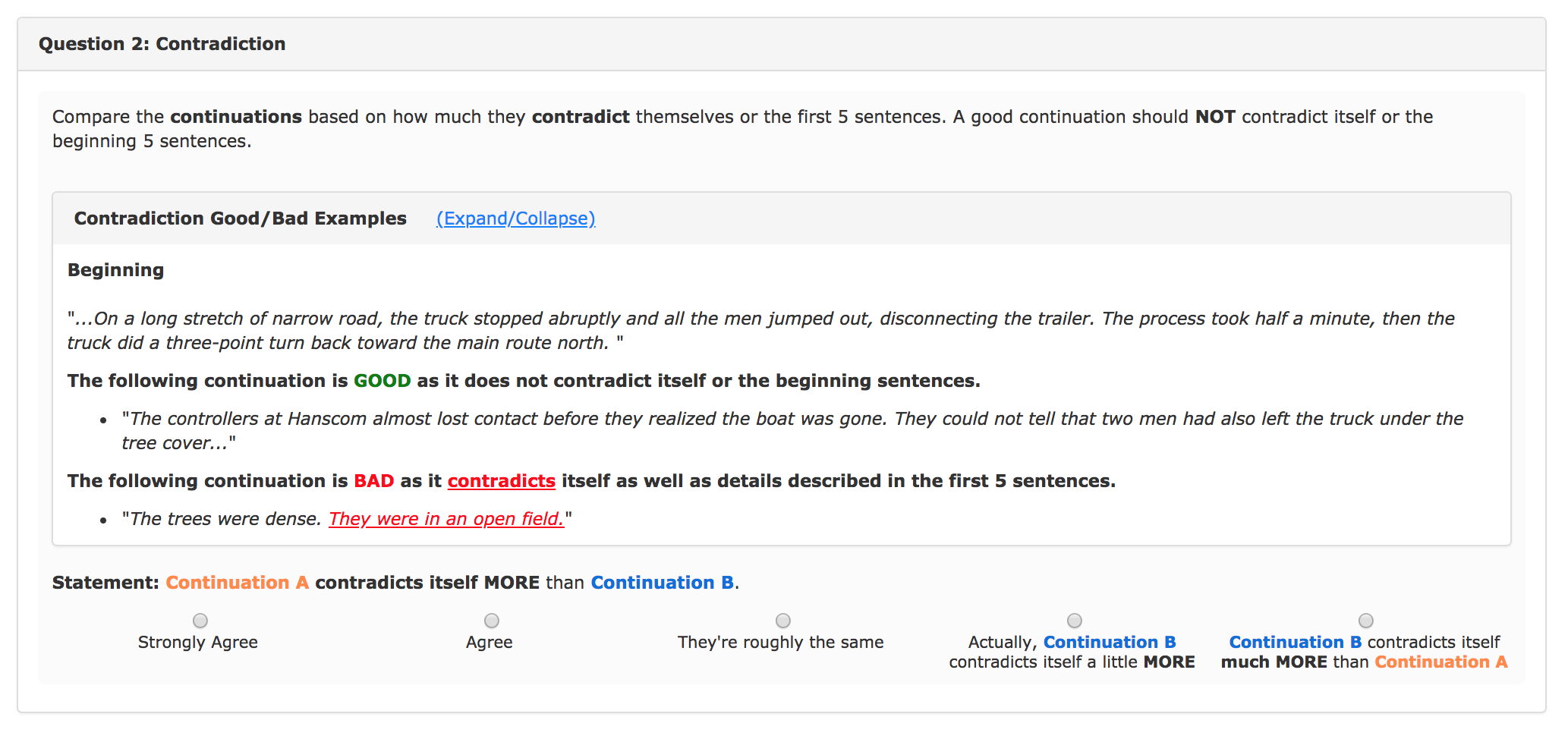}
\caption{The first half of the form for the BookCorpus human evaluation.}
\label{table:tbooks1}
\end{table*}

\begin{table*}
\includegraphics[width=6in]{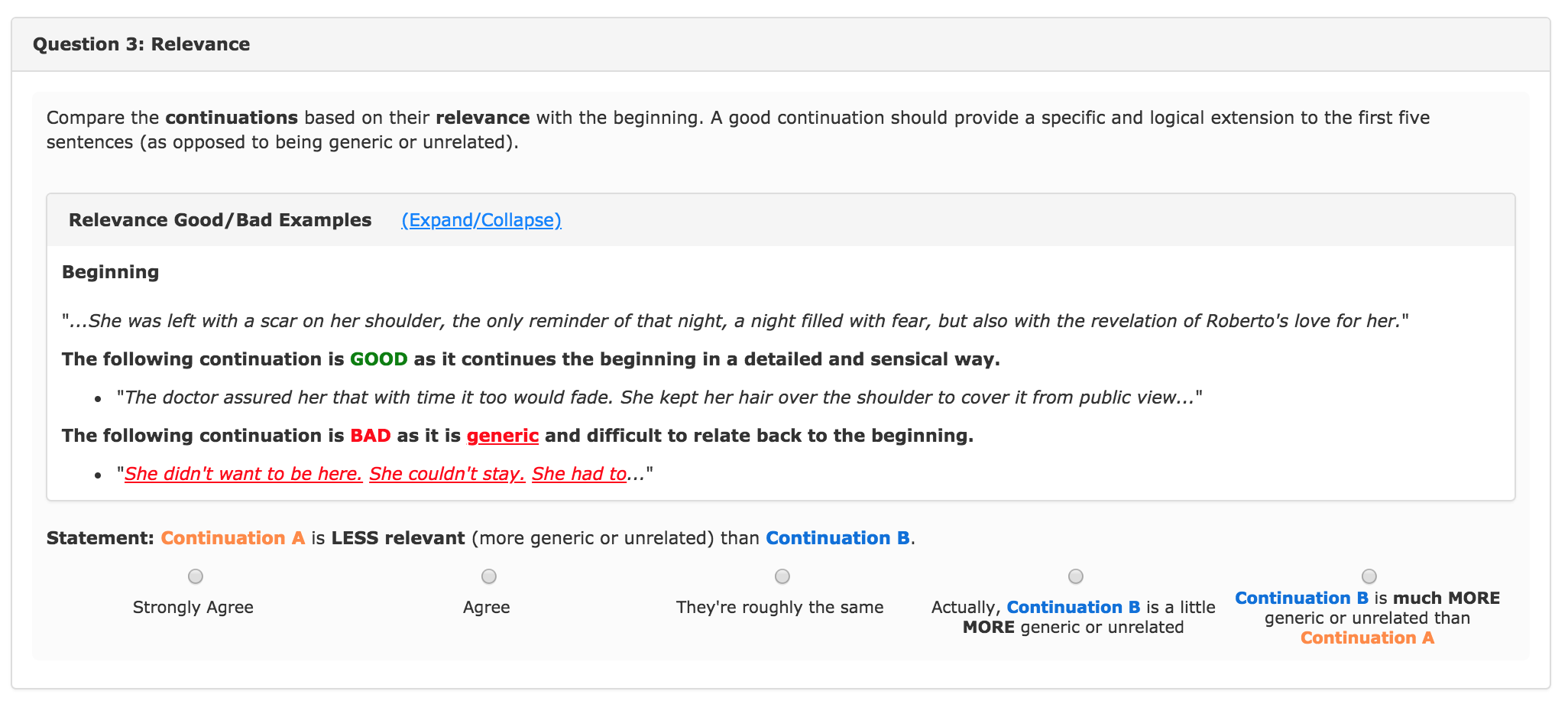}
\includegraphics[width=6in]{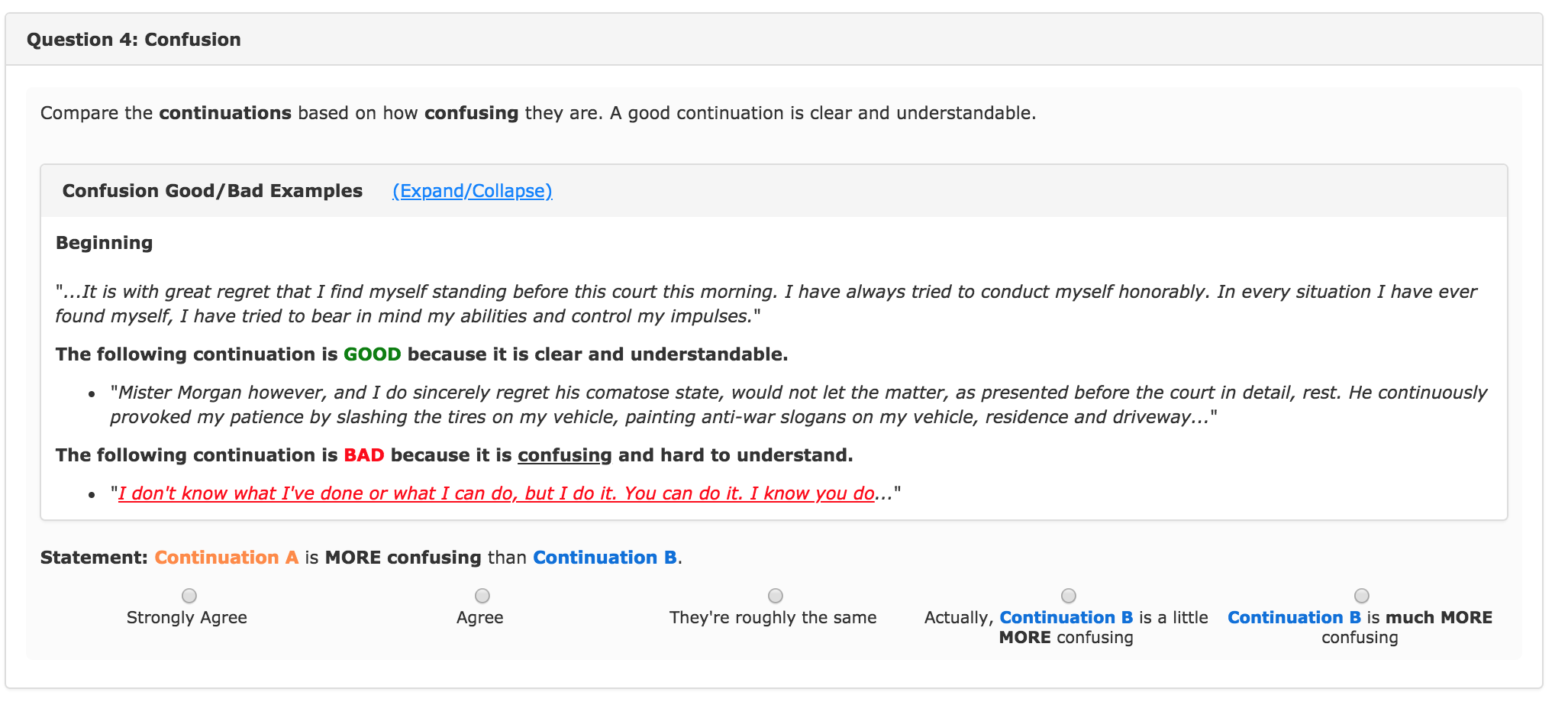}
\includegraphics[width=6in]{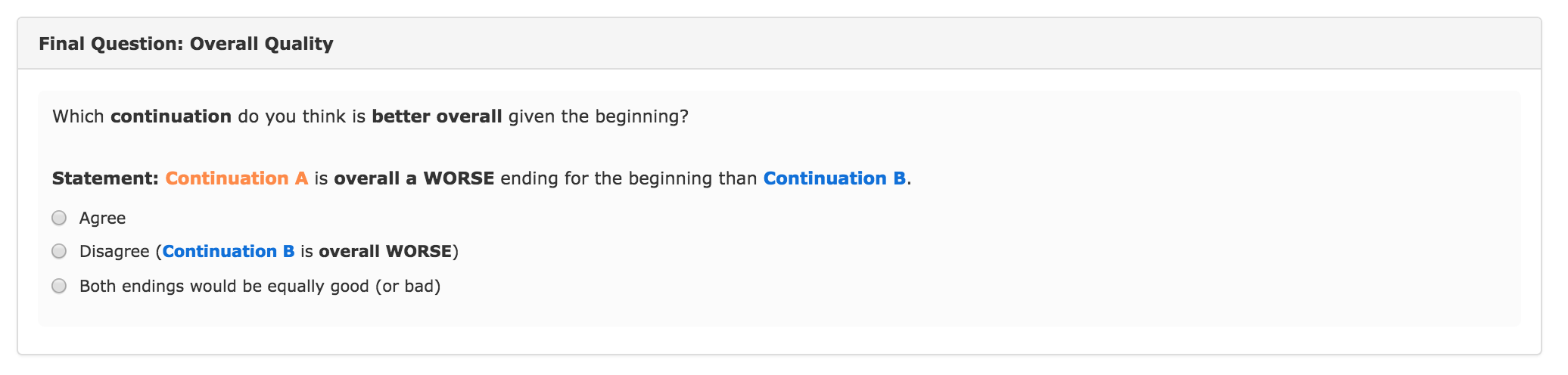}
\includegraphics[width=6in]{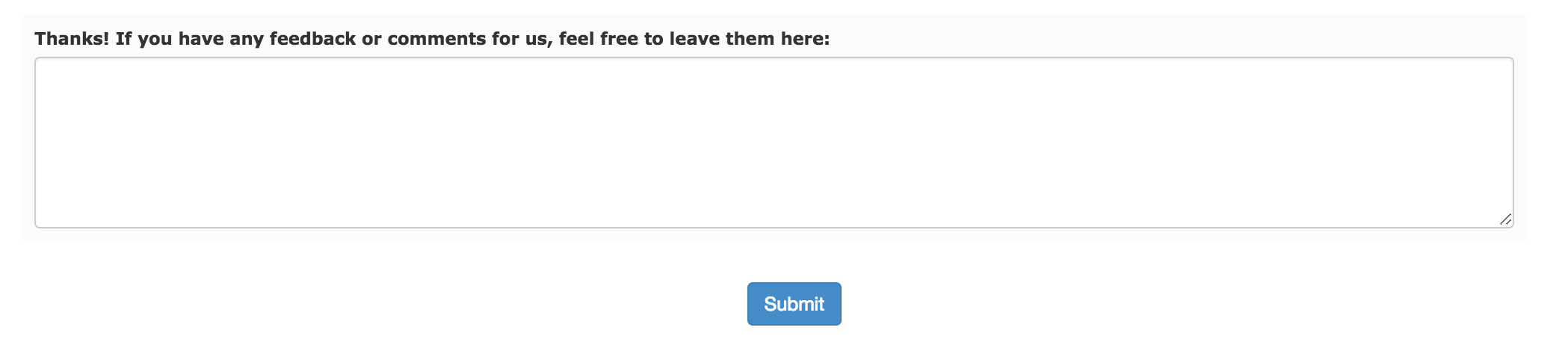}
\caption{The second half of the form for the BookCorpus human evaluation.}
\label{table:tbooks2}
\end{table*}

\begin{table*}
\center{\textbf{TripAdvisor}}

\includegraphics[width=6in]{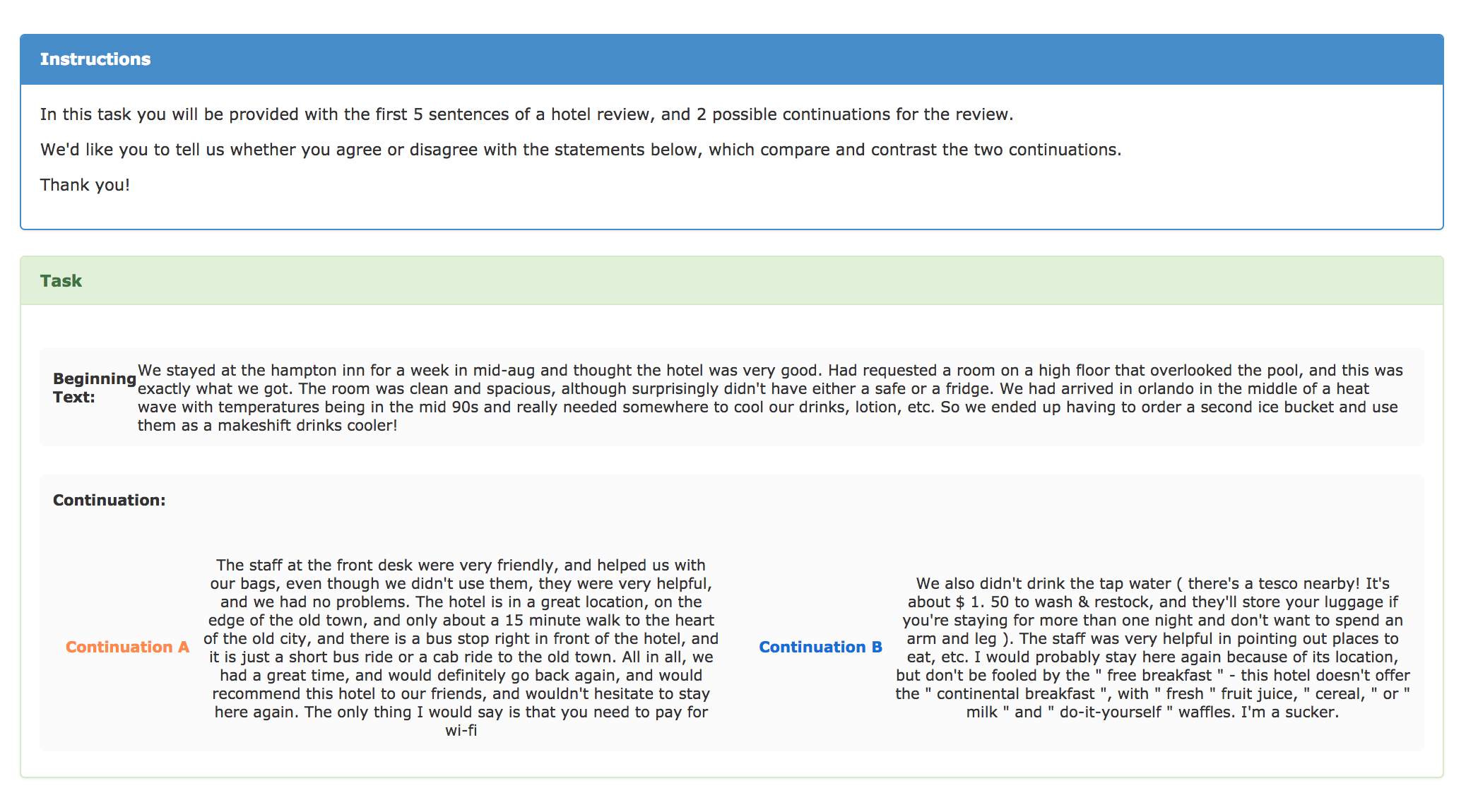}
\includegraphics[width=6in]{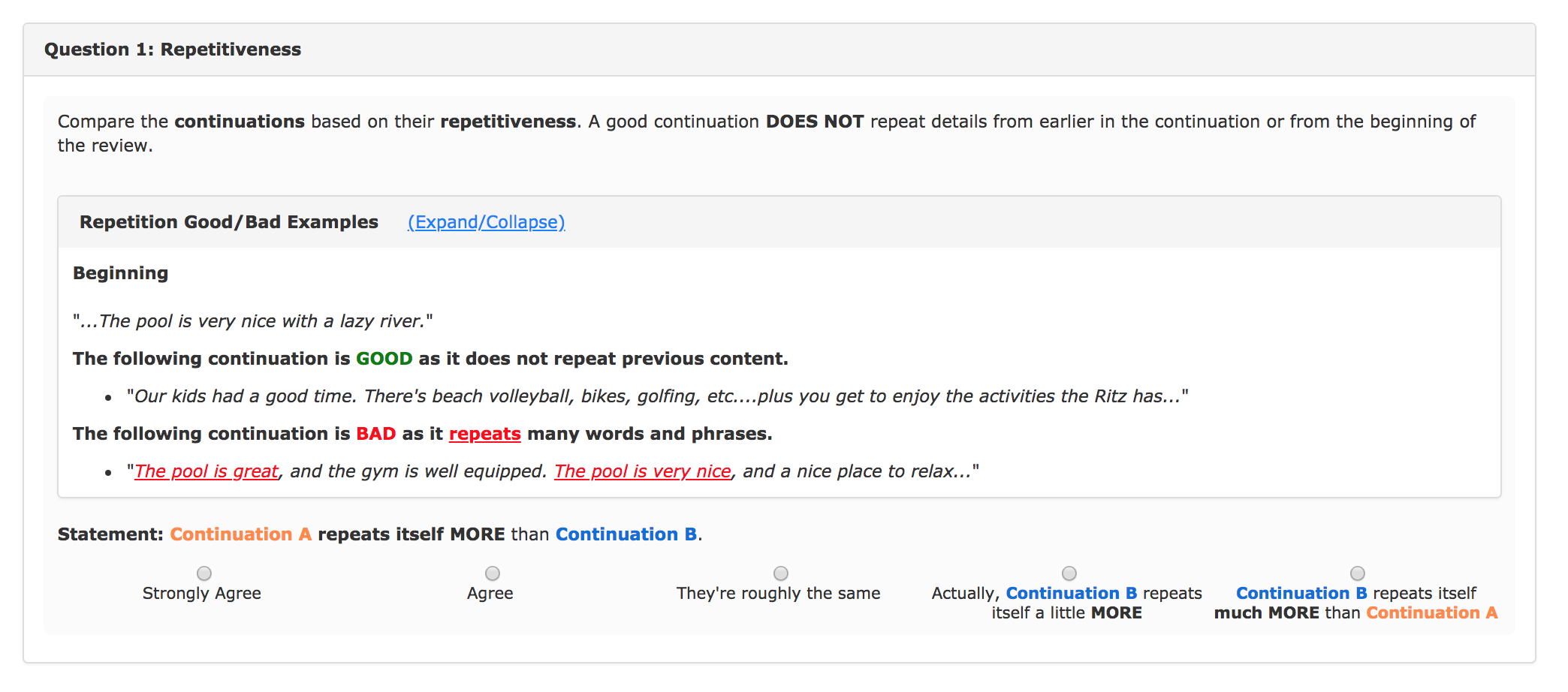}
\includegraphics[width=6in]{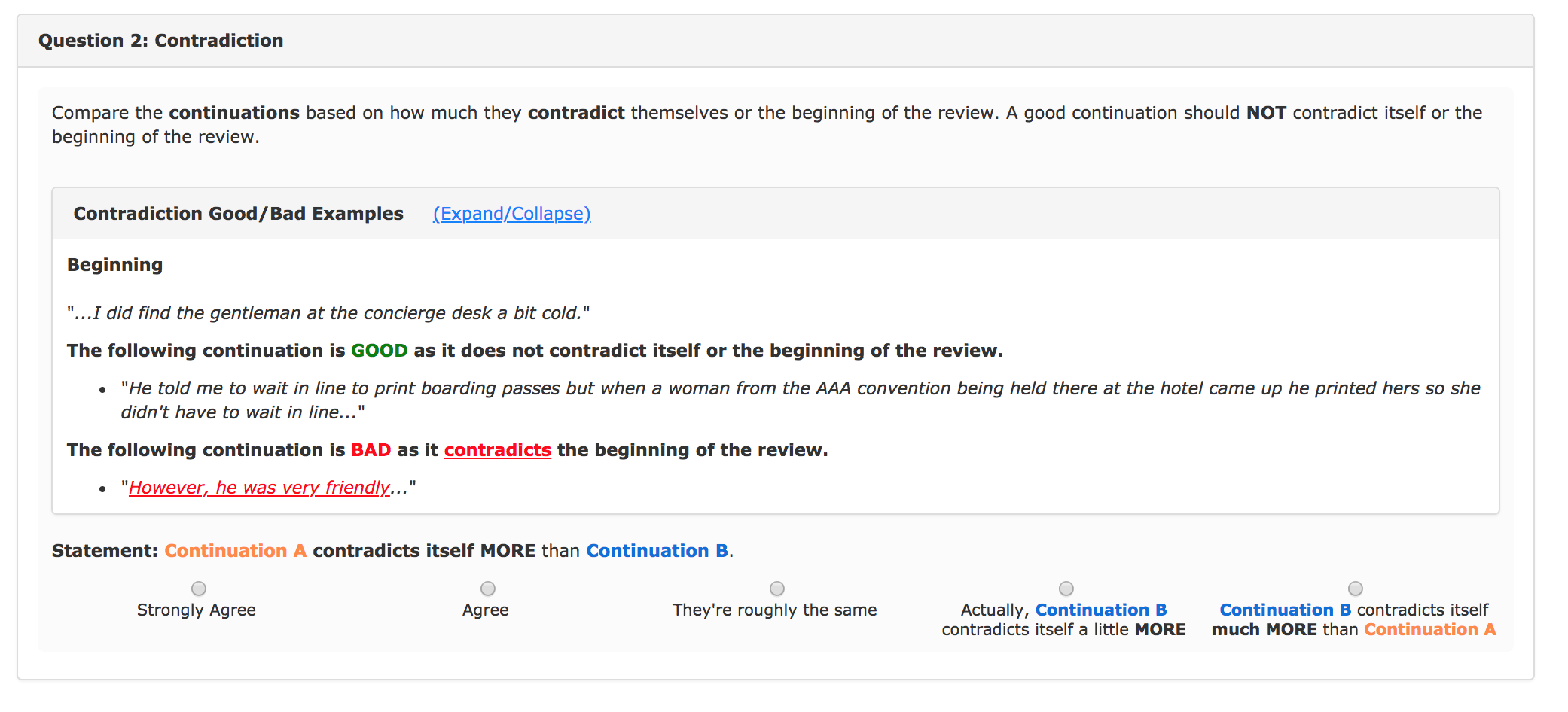}
\caption{The first half of the form for the TripAdvisor human evaluation.}
\label{table:trip1}
\end{table*}

\begin{table*}
\includegraphics[width=6in]{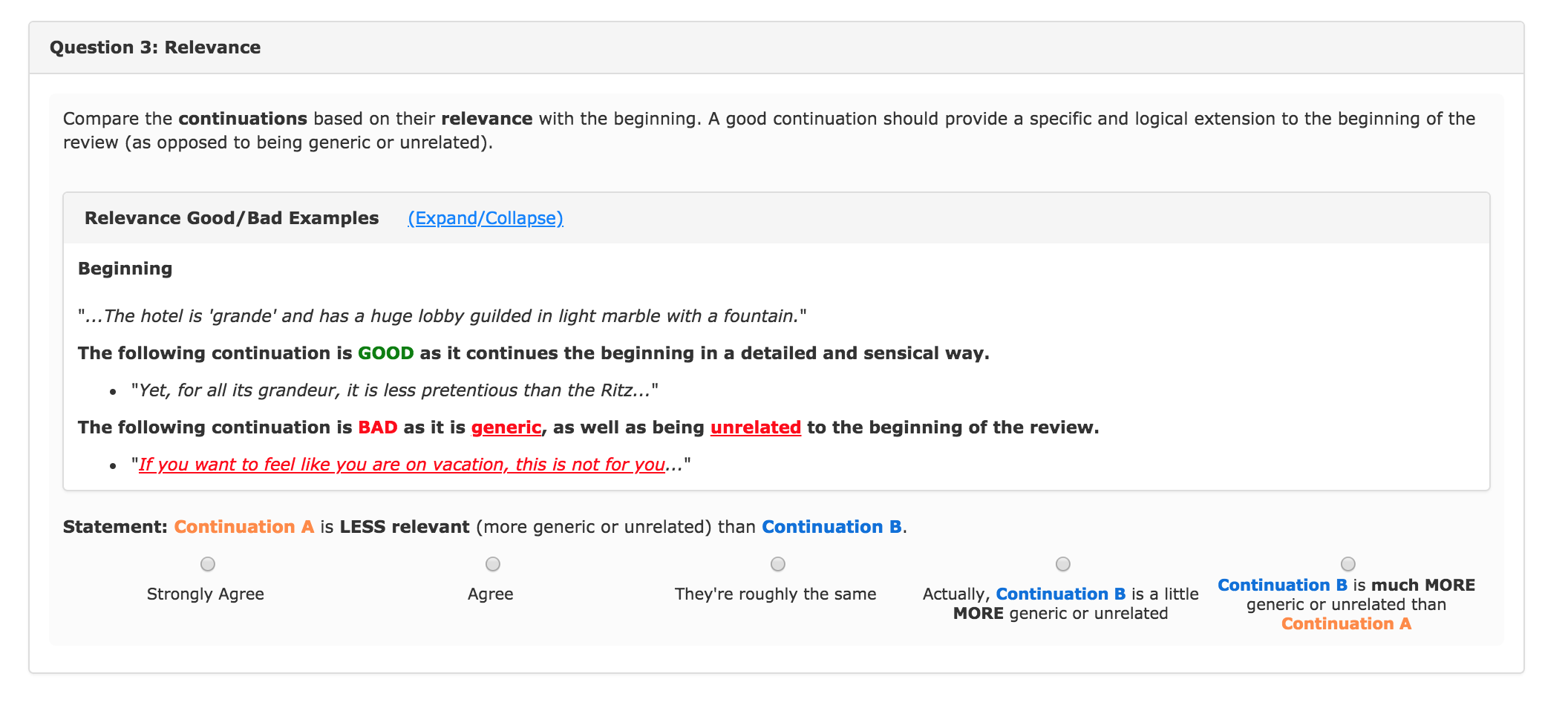}
\includegraphics[width=6in]{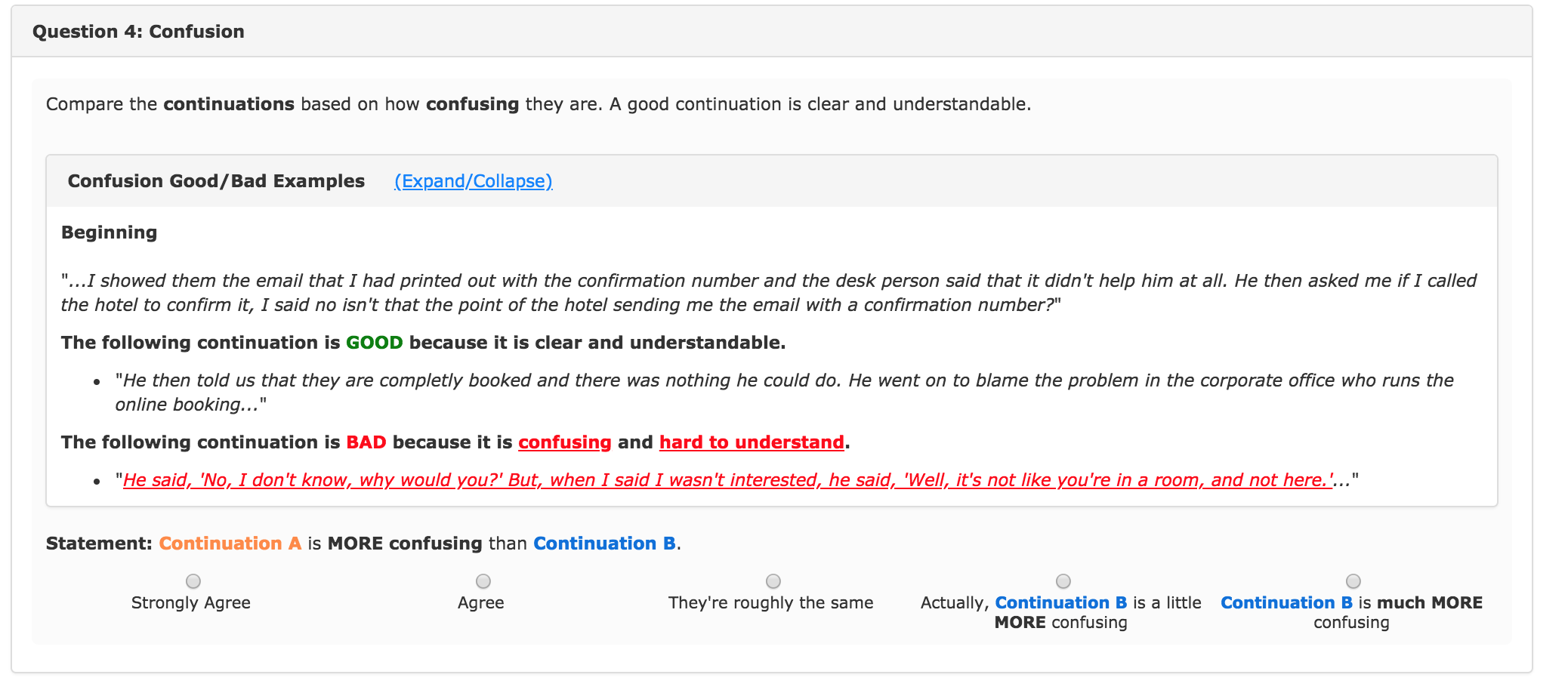}
\includegraphics[width=6in]{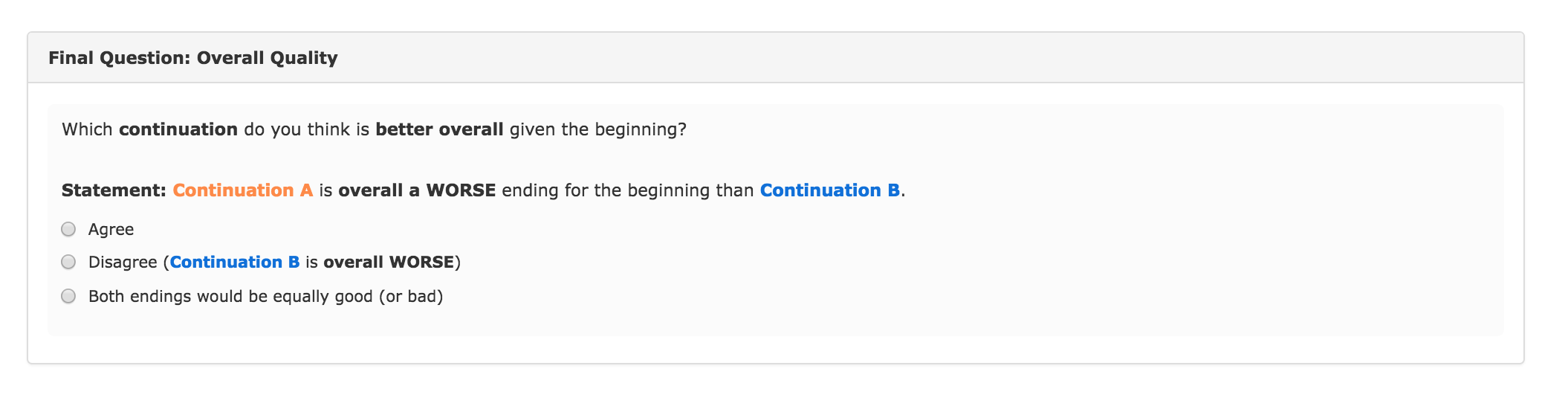}
\includegraphics[width=6in]{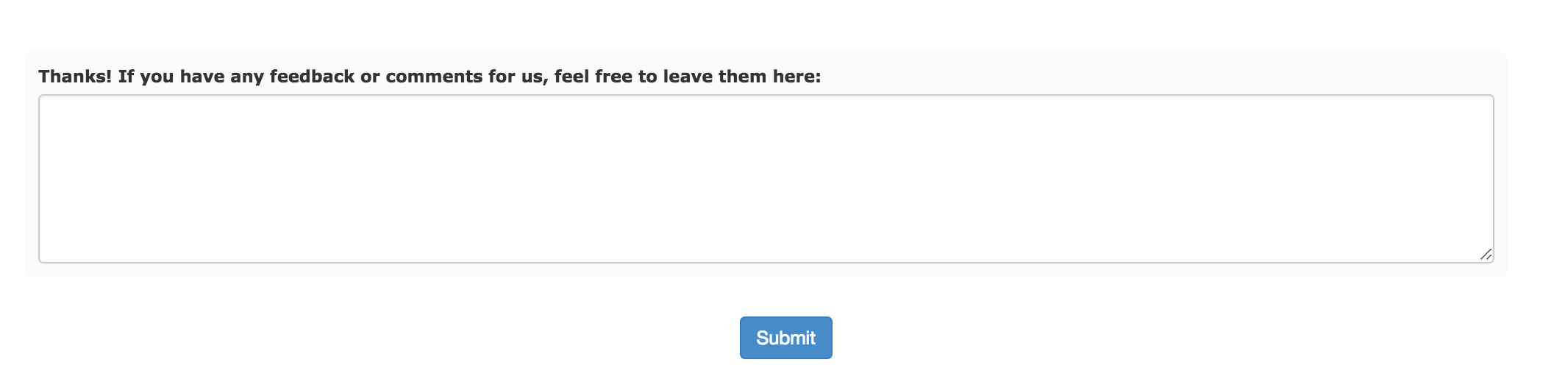}
\caption{The second half of the form for the TripAdvisor human evaluation.}
\label{table:trip2}
\end{table*}

\end{document}